# On the Partition Function and Random Maximum A-Posteriori Perturbations

Tamir Hazan    TAMIR@TTIC.EDU
Tommi Jaakkola    TOMMI@CSAIL.MIT.EDU

## Abstract

In this paper we relate the partition function to the max-statistics of random variables. In particular, we provide a novel framework for approximating and bounding the partition function using MAP inference on randomly perturbed models. As a result, we can use efficient MAP solvers such as graph-cuts to evaluate the corresponding partition function. We show that our method excels in the typical "high signal - high coupling" regime that results in ragged energy landscapes difficult for alternative approaches.

## 1. Introduction

Learning and inference in complex models drives much of the research in machine learning applications, from computer vision, natural language processing, to computational biology. Examples include object detection (Felzenszwalb et al., 2009), stereo vision (Szeliski et al., 2007), parsing (Koo et al., 2010), or protein design (Sontag et al., 2008). The inference problem in such cases involves assessing the likelihood of possible structures, whether objects, parsers, or molecular structures. The structures are specified by assignments of random variables that need to be maximized or summed over. However, it is often feasible to only find the most likely or maximum a-posteriori (MAP) assignment rather than considering all possible assignments. Indeed, substantial effort has gone into developing algorithms for recovering MAP assignments, either based on specific structural restrictions such as super-modularity (Kolmogorov, 2006) or by devising approximate methods based on linear programming relaxations (Sontag et al., 2008; Werner, 2008).

MAP inference is limited when there are other likely assignments. For example, in pose estimation the recovery of the 3D joint positions from 2D images is often inherently ambiguous. Similarly, in parsing there might be equally likely parse trees for the same sentence. In a fully probabilistic treatment, all possible alternative assignments are considered. This requires summing over the assignments with their respective weights – evaluating the partition function – which is considerably harder. In fact, any algorithm for computing the partition function would be able to approximate the MAP value arbitrarily well via a temperature argument (Landau & Lifshitz, 1980). In contrast, MAP inference (maximization) can be tractable even when the problem of evaluating the partition function (weighted counting) is not.

The main surprising result of our work is that MAP inference can be used to approximate the partition function. In other words, given an algorithm for computing the MAP value, we can approximate or bound the partition function. The MAP values in our case arise from randomly perturbed models. While models based on random perturbations have been considered recently (Papandreou & Yuille, 2011; Keshet et al., 2011; Tarlow et al., 2012), their relation to the partition function has not. Specifically, we relate the partition function to the expected MAP value of perturbations. This result enables us for the first time to directly use efficient MAP solvers such as graph-cuts or MPLP in calculating the partition function. The approach excels in regimes where there are several but not exponentially many prominent assignments. For example, this happens in cases where observations carry strong signals (local evidence) but are also guided by strong consistency constraints (couplings).

We begin by introducing the notation and the counting and maximization problems of interest. We subsequently relate the partition function to the max-statistics of random variables, and introduce new approximations and bounds on the partition function based on random MAP perturbations. Finally, we describe how to use this method in the context of conditional random fields and demonstrate the effectiveness of the approach.

## 2. Background

Here we briefly define the counting and maximization problems of interest. Throughout the paper we assume real valued potentials $\phi(y) = \phi(y_1, ..., y_n) < \infty$ defined over a discrete product space $Y = Y_1 \times \cdots \times Y_n$. The domain is implicitly defined through $\phi(y)$ via exclusions $\phi(y) = -\infty$ whenever $y \notin dom(\phi)$.

The Gibbs distribution maps the real valued potential functions to the probability scale

$$p(y_1, ..., y_n) = \frac{1}{Z} \exp(\phi(y_1, ..., y_n)) \quad (1)$$
$$Z = \sum_{y_1,...,y_n} \exp(\phi(y_1, ..., y_n)). \quad (2)$$

where the normalization constant $Z$ is also known as the partition function. The feasibility of using such a distribution for inference and learning is inherently tied to the ability to evaluate the partition function. In the special case, where $\phi(y) \in \{-\infty, 0\}$ the partition function reduces to counting the number of allowed configurations, namely $|dom(\phi)|$. In general, counting problems are considered very hard, many belonging to the complexity class #P (Valiant, 1979).

We can also express the maximum a-posteriori (MAP) inference problem in the same notation as

$$\text{(MAP)} \quad \max_{y_1,...,y_n} \phi(y_1, ..., y_n). \quad (3)$$

where, again, the domain of $\phi(y)$ is implicitly considered since the maximization avoids all configurations outside the domain. Although the MAP problem is NP-hard in general (Shimony, 1994), it is easier than computing the partition function. It can be solved efficiently in many cases of practical interest, e.g. when $\phi(y)$ is a super-modular function. A number of algorithms based on linear programming relaxations have recently been developed for solving MAP problems. Although the run-time of these solvers can be exponential, they are often surprisingly effective in practice.

## 3. Max-Statistics

In the following we describe the basis of our framework. We show how to realize the partition function as the expected value of random MAP perturbations. Analytic expressions for the statistics of a random MAP perturbation can be derived for general discrete sets, whenever independent and identically distributed random perturbations are applied for every assignment $y \in Y$. Let $\{\gamma(y)\}_{y \in Y}$ be a collection of random variables. Assume these random variables are independent and identically distributed with $F(t)$ as their cumulative distribution function, i.e. $F(t) = P[\gamma(y) \leq t]$ for each $y \in Y$. The independence of $\gamma(y)$ across $y \in Y$ implies that the cumulative distribution function of the random MAP perturbation $\max_{y \in Y} \{\phi(y) + \gamma(y)\}$ is the product of cumulative distribution functions of the individual perturbations $\phi(y) + \gamma(y)$. Therefore $P[\max_{y \in Y} \{\phi(y) + \gamma(y)\} \leq t]$ equals to $\prod_{y \in Y} F(t - \phi(y))$. Below we describe the max-stability of the Gumbel distribution and its expect value.

**Lemma 1.** *Let $\{\gamma(y)\}_{y \in Y}$ be a collection of independent random variables $\gamma(y)$ indexed by $y \in Y$, each following the Gumbel distribution whose cumulative distribution function is $F(t) = \exp(-\exp(-(t+c)))$, where $c$ is the Euler constant. Then the random variable $\max_{y \in Y} \{\phi(y) + \gamma(y)\}$ is distributed according to the Gumbel distribution and its expected value is the logarithm of the partition function:*

$$\log Z = E_\gamma \Big[ \max_{y \in Y} \{\phi(y) + \gamma(y)\} \Big]. \quad (4)$$

**Proof:** The Gumbel cumulative distribution function is closed under multiplication, namely

$$\prod_{y \in Y} F(t - \phi(y)) = \exp(-\sum_{y \in \mathcal{Y}} \exp(-(t - \phi(y) + c)))$$
$$= \exp(-\exp(-(t+c))Z) = F(t - \log Z).$$

Therefore the random variable $\max_{y \in Y} \{\phi(y) + \gamma(y)\}$ has the Gumbel distribution whose expected value is the logarithm of the partition function. □

In general each $y = (y_1, ..., y_n)$ represents an assignment to $n$ variables. In this case the theorem suggests introducing an independent perturbation $\gamma(y)$ for each such $n$-dimensional assignment $y \in Y$. The complexity of evaluating the log-partition function in this manner would be exponential in $n$. In the following we use lower dimensional random MAP perturbations as the main tool for approximating and bounding the partition function.

## 4. Low Dimensional Perturbations

In this section, we develop efficient approximations and bounds for the partition function based on low dimensional random MAP perturbations. We commence with rewriting the previous result by exploiting the structure of the product space.

**Theorem 1.** *Let $\{\gamma_i(y_i)\}_{y_i \in Y_i, i=1,...,n}$, be a collection of independent and identically distributed (i.i.d.) random variables following the Gumbel distribution with* $F(t) = \exp(-\exp(-(t+c)))$ where $c$ is the Euler constant. Define $\gamma_i = \{\gamma_i(y_i)\}_{y_i \in Y_i}$. Then

$$\log Z = E_{\gamma_1} \max_{y_1} \cdots E_{\gamma_n} \max_{y_n} \{\phi(y) + \sum_{i=1}^{n} \gamma_i(y_i)\}.$$

**Proof:** The result follows from applying equation (4) iteratively. Intuitively, $Z = \sum_{y_1} \cdots \sum_{y_n} \exp(\phi(y))$ and equation (4) encodes the correspondence between summation and expectation-maximization, namely $\sum_{y_i} \leftrightarrow E_{\gamma_i(y_i)} \max_{y_i}$. More formally, consider the recursion $\phi_{i-1}(y_1, ..., y_{i-1}) = E_{\gamma_i} \max_{y_i} \{\phi_i(y_1, \ldots, y_i) + \gamma_i(y_i)\}$, where $\phi_n(y_1, \ldots, y_n) = \phi(y_1, \ldots, y_n)$. Equation (4) implies that for each $i$, $\phi_{i-1}(y_1, ..., y_{i-1}) = \log \sum_{y_i} \exp(\phi_i(y_1, ..., y_i))$. The rest of the proof now follows by induction. $\square$

The computational complexity of the alternating procedure is still exponential in $n$. For example, the inner iteration $E_{\gamma_n} \max_{y_n} \{\phi(y_1, .., y_n) + \gamma_n(y_n)\}$ needs to be estimated exponentially many times, i.e., for every $y_1, ..., y_{n-1}$. Thus from computational perspective the alternating formulation in Theorem 1 is just as inefficient as the formulation in equation (4). We show next how Theorem 1 can be used to derive effective bounds and approximations.

### 4.1. Upper Bounds on the Partition Function

Theorem 1 directly provides easily computable upper bounds on the log-partition function. Intuitively, these bounds correspond to moving expectations outside the maximization operations, each move resulting in an additional bound but also reducing the computational effort needed for the evaluation. For example,

$$\log Z \leq E_\gamma \max_y \{\phi(y) + \sum_i \gamma_i(y_i)\} \quad (5)$$

follows immediately from moving all the expectations in front. In this case the bound is a simple average of MAP values corresponding to models with only single node perturbations $\{\gamma_i(y_i)\}_{y_i \in Y_i, i=1,...,n}$. If the maximization over $\phi(y)$ is feasible (e.g., due to supermodularity), it will typically be feasible after such perturbations as well. The upper bound can thus be evaluated efficiently as a sample average. We generalize this basic result further below.

**Corollary 1.** *Consider a family of subsets $\alpha \in \mathcal{A}$ such that $\cup_{\alpha \in \mathcal{A}} \alpha = \{1, ..., n\}$, and let $y_\alpha$ be a set of variables $\{y_i\}_{i \in \alpha}$ restricted to the indexes in $\alpha$. Assume that the random variables $\gamma_\alpha(y_\alpha)$ are i.i.d. according to the Gumbel distribution, for every $\alpha, y_\alpha$. Then*

$$\log Z \leq E_\gamma \Big[\max_{y_1, ..., y_n} \{\phi(y) + \sum_{\alpha \in \mathcal{A}} \gamma_\alpha(y_\alpha)\}\Big].$$

**Proof:** If the subsets $\alpha$ are disjoint, then $\{y_\alpha\}_{\alpha \in \mathcal{A}}$ simply defines a partition of the variables in the model. We can therefore use equation (5) over these grouped variables. In the general case, $\alpha, \alpha' \in \mathcal{A}$ may overlap. We lift the variables $y_1, \ldots, y_n$ to a larger set $y' = \{y'_\alpha\}_{\alpha \in \mathcal{A}}$ where an independent set of variables is introduced for each $\alpha \in \mathcal{A}$. We lift the potentials to $\phi'(y')$ by including consistency constraints among the lifted variables

$$\phi'(y') = \begin{cases} \phi(y_1, ..., y_n) & \text{if } \forall \alpha, i \in \alpha : y'_{\alpha,i} = y_i \\ -\infty & otherwise \end{cases}$$

Thus, $\log Z = \sum_{y'} \exp(\phi'(y'))$ since inconsistent settings receive zero weight. Moreover, $\max_{y'} \{\phi'(y') + \sum_\alpha \gamma_\alpha(y'_\alpha)\}$ equals $\max_y \{\phi(y) + \sum_\alpha \gamma_\alpha(y_\alpha)\}$ for each realization of the perturbation. This equality holds after expectation over $\gamma$ as well. Now, given that the perturbations are independent for each lifted coordinate, the basic result in equation (5), guarantees that $E_\gamma \big[\max_{y'} \{\phi'(y') + \sum_\alpha \gamma_\alpha(y'_\alpha)\}\big]$ upper bounds $\log Z$. This completes the proof. $\square$

The structure of $\alpha \subset \{1, ..., n\}$ determines the statistical quality and algorithmic efficiency of the method. For example, using a single set $\alpha = \{1, ..., n\}$ the upper bound turns to be the exact characterization in equation (4), but it requires exponentially many independent random variables.

### 4.2. Approximating the Partition Function

We can also use Theorem 1 to derive sampling based approximation schemes for the partition function that also utilize efficient MAP solvers. As a simple step, we could just replace the expectations in Theorem 1 with sampled estimates. The main subtlety lies in how these samples are reused as part of the outer loop maximization steps.

Let's begin by considering the $n$-th dimension alone. For any given setting of $y_1, ..., y_{n-1}$, we toss random values $\gamma_n(y_n)$, for every $y_n \in Y_n$, to estimate $\log \sum_{y_n} \exp(\phi(y_1, ..., y_n))$. Since the operation has to be repeated for each different setting of $y_1, \ldots, y_{n-1}$, we need $m \cdot |Y|$ random variables $\gamma_{n,j}(y_1, ..., y_{n-1}, y_n)$, for every $j = 1, ..., m$, to ensure that

$$\frac{1}{m} \sum_{j=1}^{m} \max_{y_n} \{\phi(y_1, ..., y_n) + \gamma_{n,j}(y_1, ..., y_n)\}.$$

approximates $\log \sum_{y_n} \exp(\phi(y_1, ..., y_n))$ across all $y_1, \ldots, y_{n-1}$. We can rewrite this expression in terms of a single maximization problem over an extended set of variables. Specifically, we introduce copies $y_{n,j}$ for each sample index $j = 1, \ldots, m$, and pull the maximizations outside the average:

$$\max_{y_{n,1},\ldots,y_{n,m}} \frac{1}{m} \sum_{j=1}^m (\phi(y_1, ..., y_{n,j}) + \gamma_{n,j}(y_1, ..., y_{n,j})).$$

This result is an approximation to the last expectation-maximization step in Theorem 1. We can now repeat the procedure for dimension $n-1$. Formally, we are using the fact that the partition function is self-reducible, i.e., $\log Z = \log \sum_{y_1,\ldots,y_{n-1}} \exp(\log \sum_{y_n} \exp(\phi(y)))$. After repeating this step for all the dimensions, we write the resulting approximation as a *single* but very large MAP program:

$$\log Z \approx \max_y \quad \frac{1}{m^n} \sum_{j_1,\ldots,j_n=1}^m \phi(y_{1,j_1}, ..., y_{n,j_n}) + \sum_{i=1}^n \frac{1}{m^i} \sum_{j_1,\ldots,j_i=1}^m \gamma_{i,j_i}(y_{1,j_1}, ..., y_{i,j_i}).$$

In this expression, the maximization is over an inflated set of variables where there are $m$ copies of each variable, e.g., $y_i$ is expanded into $\{y_{i,j_i}\}_{j_i=1,\ldots,m}$. The problem is that this program uses exponentially many independent random variables $\gamma_{i,j_i}(y_{1,i_1}, ..., y_{i,j_i})$ and thus suffers from the same computational problem as equation (4). However, due to averaging, we expect that most MAP perturbations are now concentrated around the logarithm of the partition function.

For computational efficiency we suggest reusing the perturbations, collapsing independent random variables $\gamma_{i,j_i}(y_{1,i_1}, ..., y_{i,j_i})$ to far fewer random variables $\gamma_{i,j_i}(y_{i,j_i})$. This results in the following approximation for the log-partition function:

$$\max_{y_{1,j_1},\ldots,y_{n,j_n}} \frac{1}{m^n} \sum_{j_1,\ldots,j_n} \phi(y_{1,j_1}, ..., y_{n,j_n}) + \sum_{i,j_i} \gamma_{i,j_i}(y_{i,j_i})$$

This approximation is effective whenever the dominant configurations of variables $\{y_{i,j_i}\}$ were already correlated (few modes) so that the randomness in $\gamma_{i,j_i}(y_{1,i_1}, ..., y_{i,j_i})$ could be compressed with little loss in accuracy. This behavior is typical in models with strong couplings.

### 4.3. Lower Bounds on the Partition Function

For completeness, we also provide a lower bound on the partition function based on randomized MAP computations. Unlike the upper bound discussed earlier, however, the lower bound does not directly exploit Theorem 1.

**Theorem 2.** *Consider any collection of subsets $\alpha \subset \{1, ..., n\}$ and let $\{\gamma_\alpha(y_\alpha)\}$ be independent random variables for each setting of $\alpha, y_\alpha$. Let $K_{\alpha,y_\alpha}(\lambda) = \log E[\exp(\lambda \gamma_\alpha(y_\alpha))]$. Then $\log Z \geq$*

$$\sup_{\lambda \geq 0} \left\{ \begin{array}{l} \log E_\gamma \left[ \exp(\max_y \{\phi(y) + \lambda \sum_\alpha \gamma_\alpha(y_\alpha)\}) \right] \\ - \max_y \sum_\alpha K_{\alpha,y_\alpha}(\lambda) \end{array} \right\}$$

The proof appears in the supplementary material. The lower bound is somewhat weaker than the upper bound in Corollary 1. For example, unlike the upper bound, in case of a single $\alpha = \{1, ..., n\}$, the lower bound is not tight for every $\phi(y)$. The parameter $\lambda$ governs the tradeoff between the perturbed MAP value and the cumulant generating function $K_{\alpha,y_\alpha}(\lambda)$. For $\lambda = 0$ the cumulant generating function is zero and the lower bound reduces to a trivial bound based on the MAP value. However, we cannot choose arbitrarily large $\lambda$ since $K_{\alpha,y_\alpha}(\lambda)$ can be unbounded depending on the distributions used for perturbations.

## 5. Conditional Random Fields

In a supervised learning problem, we assume training data $\mathcal{S}$ of objects $x \in X$ and labels $y \in Y$ such as images and their segmentations. Given a feature vector $\Phi(x, y) < \infty$ for each object $x \in \mathcal{X}$ and label $y \in Y$, the learning task is to estimate parameters $\theta$ that maximize the log likelihood of the data under the conditional random field model $p_x(y; \theta) = \exp(\theta^T \Phi(x, y))/Z_x(\theta)$. This task can be equivalently stated as a loss minimization problem

$$\min_\theta \sum_{(x,y) \in S} \left( \log Z_x(\theta) - \theta^\top \Phi(x, y) \right).$$

The formulation emphasizes the computational cost of using conditional random fields due to the partition function. In the following we make use of a surrogate partition function arising from random MAP perturbations.

Consider a family of subsets $\alpha \subset \{1, ..., n\}$ that cover $\{1, \ldots, n\}$, and let $\{\gamma_\alpha(y_\alpha)\}_{\alpha,y_\alpha}$ be i.i.d. random variables with continuous densities. We use the following surrogate criterion for estimating $\theta$: $J(\theta) =$

$$\sum_{(x,y) \in S} \left( E_\gamma[\max_{\hat{y}} \{\theta^\top \Phi(x, y) + \sum_\alpha \gamma_\alpha(y_\alpha)\}] - \theta^\top \Phi(x, y) \right)$$

The theorem below establishes some basic properties of this criterion.

**Theorem 3.** *Under the above assumptions, $J(\theta)$ is convex and smooth, and its gradient enforces the mo-*

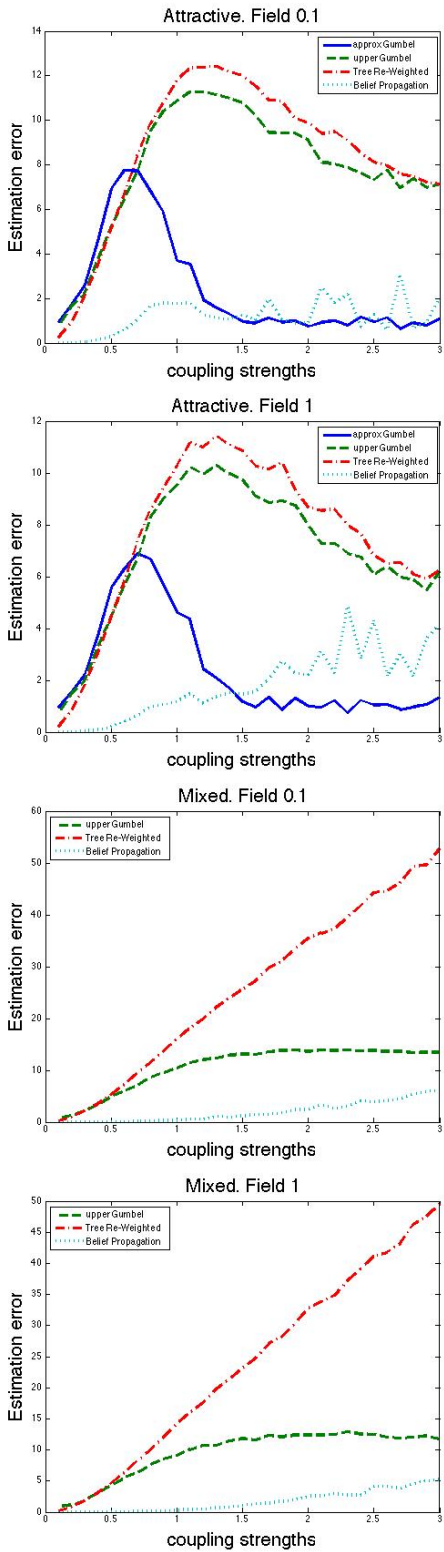

Figure 1. Comparing random MAP perturbations with tree re-weighted and belief propagation estimations for the log-partition on $10 \times 10$ spin glass model with weak and strong local field potentials. The plots describe the absolute estimation error, averaged over $100$ trials.

ment matching constraints

$$\frac{\partial}{\partial \theta} J(\theta) = \sum_{(x,y) \in S} \Big( \sum_{y'} \hat{p}_x(y') \Phi(x,y') - \Phi(x,y) \Big)$$

where

$$\hat{p}_x(y') \stackrel{def}{=} P\Big[y' \in \underset{\hat{y}_1,...,\hat{y}_n}{argmax}\{\theta^\top \Phi(x,\hat{y}) + \sum_\alpha \gamma_\alpha(\hat{y}_\alpha)\}\Big].$$

In particular, if $\gamma_\alpha(y_\alpha)$ have the Gumbel distribution, $J(\theta)$ upper bounds the conditional random field loss.

**Proof:** The loss function is convex in $\theta$, cf. (Rockafellar, 1974) Theorem 3. Corollary 1 implies that the surrogate loss upper bounds the partition-loss. To compute the gradient we use Theorem 23 in (Rockafellar, 1974) to differentiate under the integral. The subgradient of the max-function is the indicator function over the maximum argument, cf. Proposition 4.5.1 in (Bertsekas et al., 2003). Since the expectation over the indicator function results in a probability distribution, the surrogate loss is smooth and the gradient takes the above form. □

The structure of $\alpha \subset \{1,...,n\}$ determines the statistical quality and algorithmic efficiency of the approximation. For example, if we use a single set $\alpha = \{1,...,n\}$, this formulation is an exact characterization of the conditional random fields, and its gradient describes the standard moment matching condition.

## 6. Empirical Evaluation

We evaluated our approach on spin glass models

$$\phi(y_1,...,y_n) = \sum_{i \in V} \phi_i(y_i) + \sum_{(i,j) \in E} \phi_{i,j}(y_i, y_j).$$

where $y_i \in \{-1, 1\}$. Each spin has a local field parameter $\phi_i(y_i) = \theta_i y_i$ and interacts in a grid shaped graphical structure with couplings $\phi_{i,j}(y_i, y_j) = \theta_{i,j} y_i y_j$. Whenever the coupling parameters are positive the model is called attractive as adjacent variables give higher values to positively correlated configurations. We used low dimensional random perturbations $\gamma_i(y_i)$ since such perturbations do not affect the complexity of the MAP solver.

Evaluating the partition function is challenging when considering strong local field potentials and coupling strengths. The corresponding energy landscape is ragged, and characterized by a relatively small set of dominating configurations. The energy and probability landscapes are presented in the supplementary material. In the following, we show that the random MAP perturbations approach performs better than previous approaches in this setting.

We evaluated the performance of our method on $10 \times 10$ spin glass. The local field parameters $\theta_i$ were drawn uniformly at random from $[-f, f]$, where $f \in \{0.1, 1\}$ to reflect weak and strong potentials. The parameters $\theta_{i,j}$ were drawn uniformly from $[0, c]$ or $[-c, c]$ to obtain attractive or mixed coupling potentials. The following algorithms were used to estimate the partition function:

- The random MAP perturbation approximation, described in Section 4.2, was executed by inflating the graphical model to $1000 \times 1000$ grid. When dealing with attractive potentials this was efficiently evaluated using graph-cuts (Boykov et al., 2001). This approximation method could not be applied to mixed potentials.

- The random MAP perturbation upper bound, described in Section 4.1, with perturbations $\gamma_i(y_i)$. The expectation was computed using 100 random MAP perturbations, although very similar results were attained after only 10 perturbations. The MAP was computed in the attractive case using graph-cuts and in the mixed case using MPLP (Sontag et al., 2008).

- The sum-product form of tree re-weighted belief propagation with uniform distribution over the spanning trees (Wainwright et al., 2005).

- Sum-product belief propagation. Whenever the algorithm did not converge we extracted its partition function approximation from its messages.

We computed the absolute error in estimating the logarithm of the partition function, averaged over 100 spin glass models, see Fig. 1. One can see that the random MAP perturbation approximation works very well and gives a very accurate estimation using a single MAP value, when considering strong coupling potentials. Also, the random MAP perturbation upper bound is better than the tree re-weighted upper bound in the strong signal domain. Considering the attractive case, the random MAP perturbation approximation and upper bound used the graph-cuts algorithm, therefore were considerably faster than the belief propagation variants. The sum-product belief propagation performs well on the average, but from the plots one can observe its variance. This demonstrates the typical behavior of belief propagation, as it minimizes the non-convex Bethe free energy, thus works well on some instances and does not converge or attain bad local minima on others.

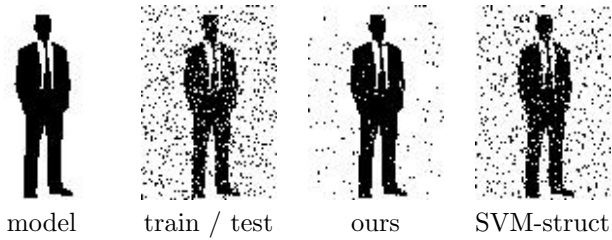

model     train / test     ours     SVM-struct

Figure 2. From left to right: (a) Binary $100 \times 70$ image. (b) A representative image in the training set and the test set, where $10\%$ of the pixels are randomly flipped. (c) A de-noised test image with our method: The test set error is $1.8\%$. (d) A de-noised test image with SVM-struct: The pixel base error is $8.2\%$.

We note that the lower bounds of random MAP perturbations, described in Section 4.3, are qualitatively the same as the MAP value, and are outperformed by standard techniques such as the mean-field. We omitted these experiments.

We also demonstrated the effectiveness of random MAP perturbations in supervised learning. The training data was composed of ten $100 \times 70$ binary images, consisting of a man silhouette and a random binary noise, described in Fig. 2. Each image $x$ is described by binary local features $\phi_i(x, y_i)$ which encodes if the $i$-th pixel is foreground or background, and pairwise features $\phi_{i,j}(y_i, y_i)$ which encourages adjacent features to have the same label. The goal is to estimate the parameters $\theta_i, \theta_{i,j}$ to de-noise the images. Since we are considering $100 \times 70$ images, there are about $20,000$ parameters to estimate. Conditional random fields cannot be evaluated on this problem, as the partition function cannot be computed to general graphs with many cycles. However, the MAP can be efficiently estimated using MPLP, thus we applied our approximate conditional random fields described in Section 5. Using the estimated parameters, the pixel based error on the test set was $1.8\%$. As mentioned before, this approach cannot be compared with conditional random fields. However, without perturbations this program relates to structured-SVM (cf. (Tsochantaridis et al., 2006)) and can be evaluated through MAP solvers. For this case, the pixel based error on the test set was $8.2\%$.

## 7. Related Work

Throughout this work, we estimate the partition function while computing the max-statistics of collections of random variables. We refer the interested reader to (Kotz & Nadarajah, 2000) for more comprehensive introduction to extreme value statistics.

To the best of our knowledge the expected value of Random MAP perturbations over discrete product spaces has not been extensively studied. Talagrand ((Talagrand, 1994), Proposition 4.3) was the first to use random MAP perturbations in discrete settings. However, their approach differs from ours in that their goal was to upper bound the size of $dom(\phi) \subseteq \{0,1\}^n$ using random variables with the Laplace distribution. The proof technique is based on a compression argument, and does not extend to the partition function. Restricting to $\phi(y) \in \{-\infty, 0\}$, Corollary 1 presents an alternative technique with weighted assignments. Another upper bound for $dom(\phi) \subseteq \{0,1\}^n$ was described in ((Barvinok & Samorodnitsky, 2007), Theorem 3.1). Their approach used the induction method of (Talagrand, 1995) to prove an upper bound using the logistic distribution. Our Corollary 1 provides an alternative technique for this result. They also extend their upper bound to functions of the form $\sum_{y \in dom(\phi)} \prod_{i:\, y_i = 1} q_i$, where $q_i$ are rational numbers.

In this work we also consider parameter estimation using approximate conditional random fields. While computing the gradient we obtained the known result that the Gibbs distribution can be described by the maximal argument of random MAP perturbation. This result is widely used in economics, providing a probabilistic interpretation for choices made by people among a finite set of alternatives. Specifically, the probability of choosing an alternative $P[\hat{y} \in \text{argmax}_y \{\phi(y) + \gamma(y)\}]$ follows the Gibbs distribution whenever $\gamma(y)$ are independent and distributed according to the Gumbel distribution (McFadden, 1974). This approach is computationally intractable when dealing with discrete product spaces, as it considers $n$-dimensional independent perturbations. This motivated efficient ways to approximately sample from the Gibbs distribution, through a probability distribution of the form: $P[\hat{y} \in \text{argmax}_y \{\phi(y) + \sum_\alpha \gamma_\alpha(y_\alpha)\}]$, (Papandreou & Yuille, 2011). In particular, the gradient suggested in Theorem 3 was described in (Papandreou & Yuille, 2011). For a more general class of probabilistic models that exploit efficient optimization we refer to (Tarlow et al., 2012). Whenever the perturbations occur in the feature space, random MAP perturbation models relate to PAC-Bayes generalization bounds (Keshet et al., 2011). Other surrogate probability models using computational structures appears in (Papandreou & Yuille, 2010; Kulesza & Taskar, 2010).

More broadly, methods for estimating the partition function were subject to extensive research over the past decades. Gibbs sampling, Annealed Importance Sampling and MCMC are typically used for estimating the partition function (cf. (Koller & Friedman, 2009) and references therein). These methods are slow when considering ragged energy landscapes, and their mixing time is typically exponential in $n$. In contrast, perturbed MAP operations are unaffected by ragged energy landscapes provided that the MAP is feasible.

Variational approaches have been extensively developed to efficiently estimate the partition function in large-scale problems. These are often inner-bound methods where a simpler distribution is optimized as an approximation to the posterior in a KL-divergence sense. The difficulty comes from non-convexity of the set of feasible distributions (e.g., mean field) (Jordan et al., 1999). Variational upper bounds on the other hand are convex, usually derived by replacing the entropy term with a simpler surrogate function and relaxing constraints on sufficient statistics (see, e.g., (Wainwright et al., 2005)).

## 8. Discussion

Evaluating the partition function and computing MAP assignments of variables are key sub-problems in machine learning. While it is well-known that the ability to compute the partition function also leads to a viable MAP algorithm, the reverse is not. We showed here that a randomly perturbed MAP solver can approximate and bound the partition function. The result enables us to take advantage of efficient MAP solvers. Moreover, we demonstrated the effectiveness of our approach in the "high-signal high-coupling" regime which dominates machine learning applications and is traditionally hard for current methods. We also applied our approximation to conditional random fields, describing the objective function to the moment matching algorithm of (Papandreou & Yuille, 2011).

The bounds we presented hold with expectation. In practice we compute the empirical mean, and standard techniques in measure concentration, e.g. Chebyshev's inequality, describe how the sampled mean relates to the expected value.

The results here can be taken in a number of different directions. The surrogate probability model, emerging from Theorem 3, is based on the maximal argument of perturbed MAP program. This surrogate probability model directly measures the robustness of prediction. Our approach also suggests a new entropy approximation, which can be derived as the conjugate-dual of the surrogate log-partition function. This entropy approximation is used with a valid set of probability distributions induced by perturbations. This is in contrast to current approximations, e.g. based on Bethe entropies, defined over the local marginal polytope.